\documentclass[runningheads]{waica}
\titlerunning{From Stateless to Situated}
\usepackage[utf8]{inputenc}
\usepackage[T1]{fontenc}
\usepackage{graphicx}
\usepackage{float}
\usepackage{booktabs}
\usepackage{array}
\usepackage{newunicodechar}

\newunicodechar{，}{,}

\begin{document}

\title{From Stateless to Situated: Building a Psychological World for LLM-Based Agents}
\author{
Boning Zhao\inst{1}\thanks{ Boning Zhao and Yutong Hu contributed equally to this work.}
\and
Yutong Hu\inst{2}\textsuperscript{$\star$}
\and
Xinnuo Li\inst{2}
}

\authorrunning{B. Zhao et al.}

\institute{
Independent Researcher, Shanghai, China\\
\email{bz2518@nyu.edu}
\and
New York University, New York, NY, USA\\
\email{\{yh4872, xl5454\}@nyu.edu}
}
\maketitle

\begin{abstract}
In long-horizon interactions, the reliance of Large Language Models (LLMs) on local next-token prediction inherently lacks the executive control required for boundary-sensitive tasks, often resulting in stateless drift. To address this fundamental limitation, we propose LEKIA 2.0, a cognitive-inspired hybrid agent architecture that explicitly decouples situational awareness from intervention execution. The Cognitive Layer maintains a continuously updatable representation of the user's internal state, while the Executive Layer enforces high-level decision constraints through a dual-gated state machine, effectively suppressing autoregressive generative impulses. To rigorously evaluate this architecture's process-control capabilities, we utilize Cognitive Behavioral Therapy (CBT)—a domain demanding strict stage progression and boundary maintenance—as an empirical testbed. Furthermore, we introduce a Static-to-Dynamic evaluation protocol to activate static corpora into interactive adversarial environments. Results demonstrate that LEKIA achieves an average absolute improvement of approximately 31\% over prompt-only baselines across deep intervention stages. Under extreme resistance, the controller produced zero cooldown violations in our stress-test setting. These findings suggest that explicitly decoupled cognitive structures can improve process-control reliability in situated agent systems under controlled multi-turn simulation.

\end{abstract}

\section{Introduction}

In the pursuit of complex autonomous agents, Large Language Models (LLMs) have demonstrated 
remarkable natural language capabilities~\cite{brown2020gpt3,openai2023gpt4}. However, their 
deployment in long-horizon, process-oriented tasks reveals a fundamental architectural limitation. 
Because standard LLMs operate primarily through local next-token 
prediction~\cite{radford2019language}, they function largely as reactive generators. Systems 
lacking an explicit external state structure often struggle to maintain a stable situational 
representation that can be updated and used to regulate complex 
processes~\cite{mrksic2017neural,heck2020trippy}. In continuous interaction, this absence of 
cross-turn situational anchors leaves generative models highly susceptible to what we term 
\textit{stateless drift}---the progressive loss of coherent situational grounding across turns.

This limitation becomes acutely problematic in environments demanding strict executive control 
and boundary maintenance. This stateless drift is further exacerbated by RLHF alignment 
biases~\cite{ouyang2022instructgpt,bai2022constitutional}: without external process control, 
generative models tend to oscillate between two extremes---either defaulting to rigid therapeutic 
didacticism (e.g., forcing advice prematurely) or drifting toward excessive affective 
accommodation, both of which fundamentally undermine intervention boundary consistency.

Existing methods often attempt to address this through prompt engineering or dialogue-history 
concatenation~\cite{wei2022cot,yao2023react}, which require the LLM to simultaneously model 
the interactional situation and generate natural language responses. This mixing of structured 
state and free-form text weakens historical coherence~\cite{zhong2023memorybank} and makes it 
exceptionally difficult to preserve rigid behavioral boundaries. We therefore propose 
\textbf{LEKIA 2.0}, a cognitive-inspired hybrid architecture that explicitly separates 
situational state modeling from intervention execution. The Cognitive Layer maintains stable 
representations of the user's internal state and process evidence across 
turns~\cite{park2023generative}, while the Executive Layer governs stage progression through a 
state machine and dual-gating mechanism. To rigorously evaluate these process-control 
capabilities, we adopt Cognitive Behavioral Therapy (CBT) as an empirical 
testbed~\cite{beck2011cbt}---a domain demanding disciplined stage control, complex stage 
transitions, and explicit resistance handling~\cite{liu2021esconv}---and introduce a 
\textbf{Static-to-Dynamic} evaluation protocol that activates static corpora into interactive 
adversarial environments via an LLM-as-User simulator~\cite{zheng2023judging}.

The main contributions are threefold. \textbf{(1)}~We propose LEKIA 2.0, a situated LLM 
architecture that explicitly separates the Cognitive Layer from the Executive Layer, using 
external state modeling and dual-gating execution to mitigate stateless drift in process-oriented 
emotional support. \textbf{(2)}~We introduce a Static-to-Dynamic evaluation protocol that 
activates static corpora into interactive multi-turn environments, providing a reusable framework 
for evaluating process-control capabilities that static text alone cannot capture. \textbf{(3)}~We 
show that LEKIA achieves substantial advantages in deep-stage completion, boundary 
maintenance under extreme resistance, and cross-scenario robustness across 11 distress types, 
while also showing extensibility to embodied virtual agent.

\section{Related Work}
\subsection{LLM Agents and Cognitive Architectures}
In the pursuit of autonomous LLM agents, longitudinal state representation has emerged as a 
fundamental challenge~\cite{wang2023survey}. Recent works have extensively explored memory 
integration and persona modeling to enhance the cross-session consistency of generative 
agents~\cite{park2023generative,zhong2023memorybank}. However, these approaches predominantly 
treat memory as an information retrieval bank---storing and fetching past dialogue snippets. In 
complex, process-oriented tasks, agents require more than historical retrieval; they need a 
continuously updatable representation of the user's narrative state and interactive 
boundaries~\cite{lewis2020rag}. Our architecture's cognitive layer departs from simple 
fact-retrieval databases by maintaining a situated, structural anchor that actively drives 
process control.

\subsection{State Tracking and Executive Control}
Dialogue State Tracking (DST) has historically been central to task-oriented systems for 
maintaining cross-turn structural consistency~\cite{mrksic2017neural,heck2020trippy}. With the 
advent of LLMs, the DST paradigm has expanded to encompass open-domain state tracking, zero-shot 
function calling, and common-ground alignment~\cite{rastogi2020towards,wu2019trade}. While these 
advancements highlight the necessity of explicit state representations beyond single-turn 
generation, they primarily target static information tracking or functional task decomposition. 
Conversely, high-stakes interactive environments demand strict \textit{executive control}: 
assessing user readiness, gating stage transitions, and managing re-offer cooldowns after 
rejection~\cite{hill2009helping,beck2011cbt}. We extend the traditional DST paradigm by 
introducing a dual-gating mechanism, which explicitly manages stage discipline and boundary preservation in complex scenarios.

\subsection{Process-Oriented Tasks as Clinical Testbeds}
To evaluate the efficacy of cognitive architectures and executive control mechanisms, 
process-oriented tasks---such as structured psychological support—serve as rigorous process-control testbeds. Early foundational work, including ESConv~\cite{liu2021esconv}, formulated emotional 
support primarily as a strategy-driven response generation problem. Subsequent research has 
significantly improved empathetic generation and strategy selection in LLM-based 
agents~\cite{qian2023harnessing,rashkin2019empathetic}. Nevertheless, these approaches 
predominantly frame the challenge as generating better supportive text, largely neglecting the 
continuous process control required in multi-turn interventions. A robust system must know when 
to explore, when to psychoeducate, and crucially, how to maintain boundaries without forcing 
premature transitions in the face of user resistance~\cite{beck2011cbt,hill2009helping}. By 
shifting the focus from single-turn empathetic generation to stable, process-oriented 
progression, this testbed exposes the critical vulnerabilities of purely generative 
models~\cite{brown2020gpt3,ouyang2022instructgpt}.

\subsection{Cognitive-Driven Embodied Interfaces}
In complex affective interactions, non-verbal signals are vital for establishing therapeutic 
alliances~\cite{knapp2013nonverbal}. While recent multimodal dialogue systems increasingly 
emphasize embodied interfaces~\cite{cassell2000embodied}, these visual representations risk 
becoming arbitrary visual decorations if they lack stable internal 
states~\cite{bickmore2005establishing}. Because our architecture explicitly decouples internal 
state from external expression, it extends naturally to multimodal embodiment. The decoupled 
design outputs structured affect instructions, ensuring that front-end embodied entities 
(e.g., Live2D) are driven by consistent internal cognitive states rather than stochastic text 
generation~\cite{park2023generative}.

\section{Methodology: Building a Situated Psychological World}

\subsection{Overall Architecture and CBT Instantiation}
To address the limitations of implicit process control in standard LLMs, LEKIA explicitly separates situational modeling from intervention execution. To maintain stable awareness of intervention stage and user boundaries throughout multi-turn interaction, LEKIA is organized into two functionally complementary layers:

Cognitive Layer. This layer serves as a situated state substrate, continuously extracting implicit cues and readiness signals from the interaction. Its role is to represent what the system currently knows about the user's situation and the evolving support context.

Executive Layer. This layer serves as a process controller, combining a macro-level stage skeleton with micro-level gating rules to determine whether to maintain the current state, issue a transition offer, or advance.

\textbf{CBT Instantiation:} Process-oriented emotional support provides a particularly demanding testbed because it requires strong intervention discipline and explicit boundary control. In this work, we instantiate the general architecture in a CBT-inspired emotional support setting. The four-stage skeleton—Assessment, Education, Intervention, and Homework—maps to Beck's CBT session structure \cite{beck2011cbt}, from initial case conceptualization through psychoeducation and cognitive restructuring to a final consolidation. Stage-specific objectives and completion criteria are detailed in Section~\ref{sec:metrics}. Throughout this progression, the system is expected to maintain intervention discipline and prevent premature advancement when the user shows resistance. The following subsections describe how the general architecture supports deep state tracking and boundary-sensitive control in this CBT instantiation.

\begin{figure*}[t]
\centering
\includegraphics[width=0.96\textwidth]{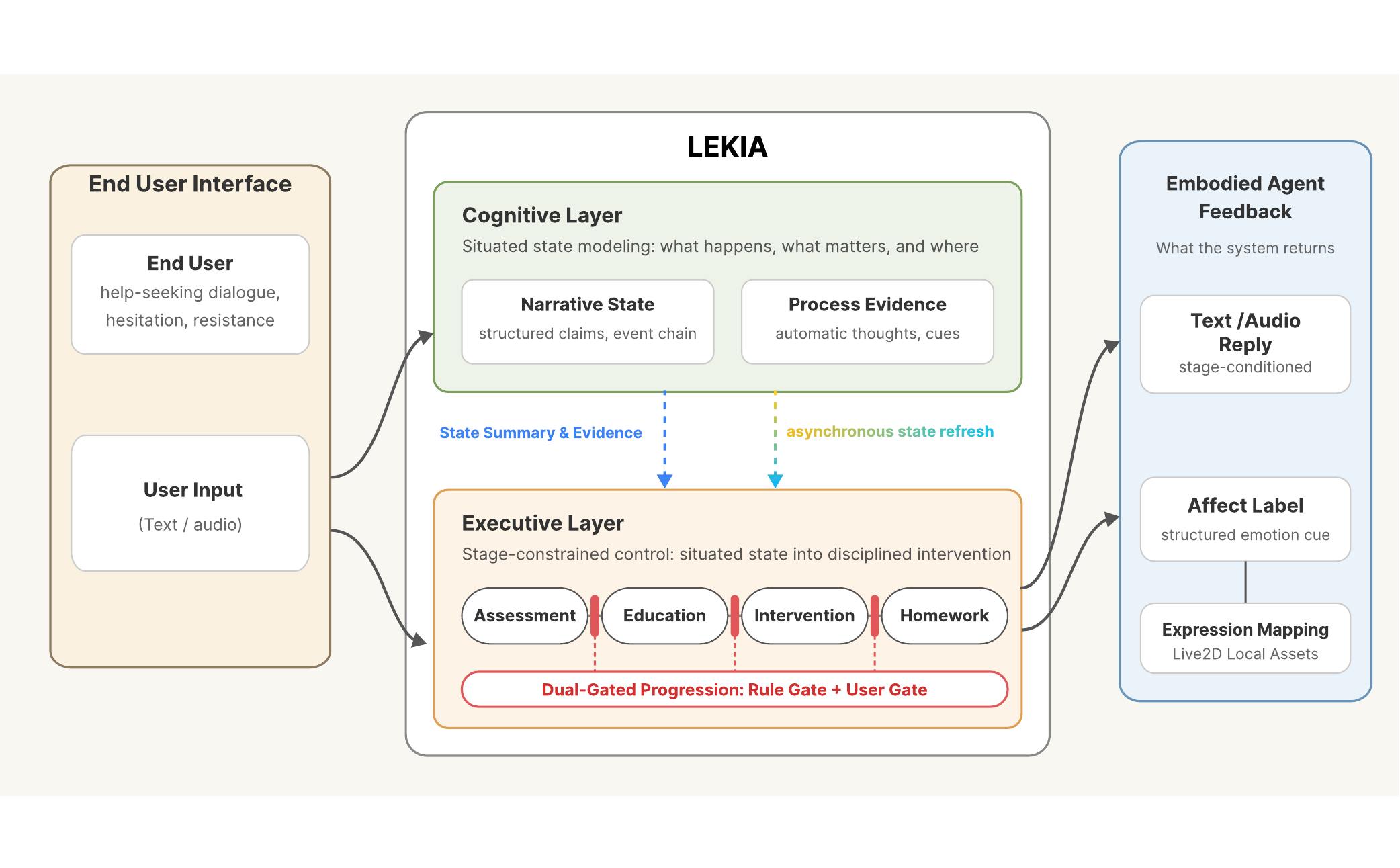}
\caption{LEKIA situated architecture.}
\label{fig:lekia-main-arch}
\end{figure*}

\subsection{Cognitive Layer: Contextual State Modeling}
The Cognitive Layer compresses ongoing interaction into a structured, continuously updateable external state. In the current CBT instantiation, it maintains two primary categories of information: Narrative State and Process Evidence.

\textbf{Narrative State}

The system continuously organizes information revealed in the dialogue into a compact narrative profile, including the user's key event chain, current conflict structure, and explicitly expressed stances. In addition, it maintains a set of structured claims: atomic narrative facts extracted from the conversation, such as interpersonal relationships, critical events, or explicitly stated attitudes. Each claim is associated with a status marker indicating whether it is currently valid, denied, or in conflict. Together, these representations provide stable long-range anchors for the system's understanding of the current interactional situation.

\textbf{Process Evidence}

Beyond narrative content, the system periodically extracts evidence relevant to CBT progression. This includes the accumulation of automatic thoughts, cues of cognitive distortion, and signs of emerging awareness or reflective readiness. These signals are not directly used for response generation. Instead, they function as external evidence used by the Executive Layer to determine whether advancement is warranted.

Both Narrative State and Process Evidence are extracted via a structured LLM call constrained by a Pydantic schema, executing asynchronously to avoid blocking the response path. Conflicting claims are resolved by recency: the most recent explicit user statement supersedes prior inferences.

\subsection{Executive Layer: Stage-Constrained Intervention and Dual-Gated Advancement}
In the CBT instantiation, the Executive Layer is organized around a four-stage process skeleton: Assessment, Education, Intervention, and Homework. These stages are not independent dialogue templates, but ordered segments within a constrained progression path. The system tracks not only the current macro-stage, but also stage-local progress flags recording whether key actions within the current stage have been completed.

For example, within the Education stage, the system distinguishes between whether a recurring pattern has been pointed out (pattern\_pointed) and whether concrete guidance has been provided (advice\_given). These local flags are critical for preventing false advancement---cases in which the response appears to move forward rhetorically while the substantive objective of the current stage has not yet been completed.

This stage-constrained design allows intervention progression to depend on explicit state tracking rather than on the apparent forward motion of text alone.

\subsection{Stage Advancement Protocol: Dual-Gating and Cooldown}
To regulate cross-stage transition, LEKIA implements an explicit Stage Advancement Protocol. This protocol separates the question of whether the system can advance from whether it should advance by introducing two sequential gates: a Rule Gate and a User Gate. The purpose of this decomposition is to formalize the kind of bounded proactivity required in process-oriented emotional support.

The Rule Gate evaluates whether the conditions for advancement have been met on the basis of process evidence from the Cognitive Layer and stage-local progress maintained by the Executive Layer. Only when these criteria are satisfied may the system issue a low-pressure Transition Offer. The User Gate then determines, based on semantic signals from the most recent user turn, whether the user is receptive to further advancement at the current pace.

This dual-gating design serves two complementary functions. First, it prevents the system from remaining trapped in ineffective empathic continuation when advancement is actually warranted. Second, it prevents advancement from being driven by turn-local generative bias rather than explicit process control. In this sense, dual-gating transforms intervention progression from an implicit tendency of language generation into an auditable, controllable system decision.

If the user refuses, hesitates, or shifts the topic, advancement is blocked and a turn-based cooldown is triggered to suppress repeated transition offers. In the current implementation, the cooldown window is set to 3 turns. Once the cooldown period expires, the Executive Layer re-evaluates the updated cognitive state before deciding whether transition inquiry should resume. This mechanism directly encodes soft-advancement principles as system-level boundary logic rather than leaving the timing of advancement to ad hoc single-turn generation.

\subsection{System Implementation}

To operationalize LEKIA in a low-latency form, we implement a hierarchical state machine using LangGraph~\cite{langgraph2024}, where CBT macro-stages are instantiated as independent nested subgraphs. Stage transitions are evaluated against a shared persistent object that accumulates narrative state and process evidence across the interaction. Crucially, model outputs at decision points are constrained to Pydantic-validated~\cite{pydantic2024} structured schemas. Instead of relying on free-form text, the model must explicitly emit completion flags (e.g., \texttt{pointed\_out\_pattern}), which function as first-class outputs to directly gate stage-local progress in the Executive Layer. This design creates a chain of symbolic checkpoints, making stage advancement explicitly conditioned on validated state rather than rhetorically inferred from free-form text.
To mitigate the prohibitive latency of synchronous full-state 
reconstruction, LEKIA decouples foreground response generation 
from background cognitive updates. Deep state extraction executes 
asynchronously, while the foreground path reads a one-turn-lagged 
state summary. This design is grounded in a psychologically motivated 
assumption: meaningful shifts in narrative state and resistance posture 
are gradual, making a one-turn lag sufficient for accurate process 
control while keeping response latency within interactive bounds. 
Furthermore, this explicit internal state directly enables coherent 
embodied expression. The Executive Layer outputs discrete affect 
labels alongside text, which deterministically drive a model-agnostic 
Live2D front-end without requiring additional generative inference 
at the presentation layer.
\begin{figure}[H]
\centering
\includegraphics[width=0.98\columnwidth]{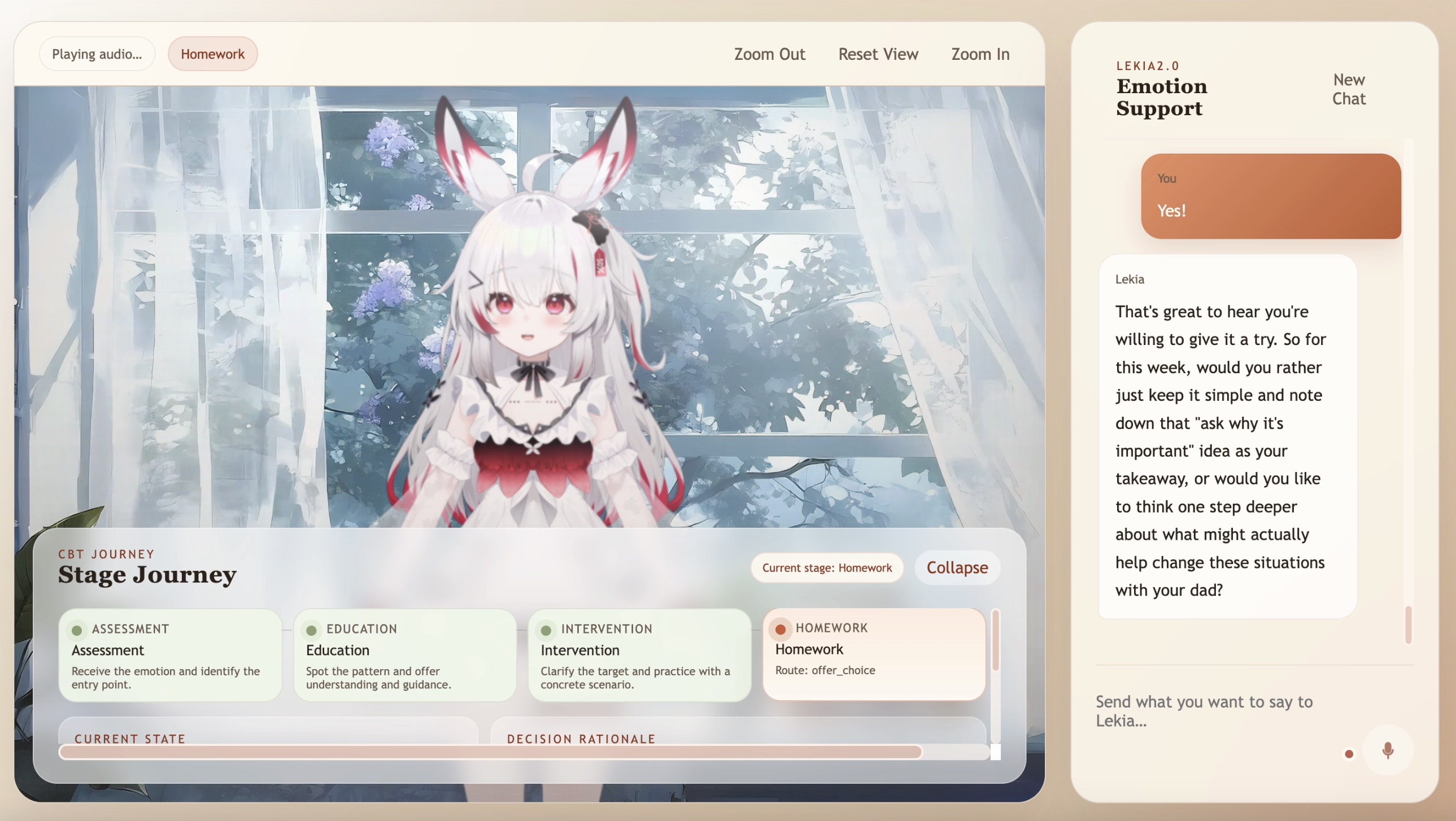}
\caption{Live2D front-end embodiment in LEKIA.}
\label{fig:live2d-frontend}
\end{figure}

\section{Experiments and Results}

\subsection{From Static Corpus to a Dynamic Evaluation Environment}
Our evaluation starting point is typical emotional support corpora such as ESConv. To adapt to the evaluation needs of process-oriented tasks, we first expanded and conducted targeted generation on the original data, constructing an expanded static case library containing 1,300 cases (covering 11 types of distress and multiple emotion labels) to endow it with longer-term intervention potential.

However, static corpora are insufficient to support the effective evaluation of dynamic Agents. If fixed dialogue texts are directly used to evaluate process-oriented systems, the system will be placed in an environment that cannot provide real-time feedback on its intervention timing, stage consultations, or boundary handling, thereby rendering the measurement of process-oriented capabilities unreliable. In the early stages of our research, we also attempted to introduce human evaluation, where human annotators played the role of seekers based on the expanded case scripts to interact with the system in real-time. Although this approach has high ecological validity, its time and labor costs are too high to support large-scale, reusable dynamic evaluations. To solve this problem, this paper proposes a Static-to-Dynamic online evaluation protocol, and on this basis, constructs a State-Blind Seeker Simulator. Anchored to the narrative settings of the original static cases, the simulator preserves the help-seeker's dilemma framing and resistance tendencies across turns, while remaining completely blind to the system's internal CBT stages, progress flags, and dual-gate status—responding solely based on the externally visible response text.

In this way, the protocol retains the static case scenario constraints while restoring the dynamic feedback properties necessary for process evaluation in long-term interactions. Based on this interactive seeker environment, we generated complete multi-turn interaction trajectories from the expanded cases and selected 300 through stratified sampling across 11 distress types, four intervention stages, and three user-response patterns for subsequent process evaluation.

\subsection{Systems and Shared Evaluation Context}
In this dynamic environment, we comparatively evaluate three types of system architectures:

LEKIA: A complete situated architecture equipped with a cognitive layer, an executive layer, explicit stage states, and a dual-gating mechanism.

Baseline: Uses a single global prompt to directly generate supportive responses, without stage divisions or explicit states.

MiddleBaseline: Adopts a strategy of hierarchical prompt switching and includes a complete prompt-based gating and cooldown mechanism, but does not maintain internal process states.

Control Variable Settings: To ensure fairness in the evaluation, the three tested systems share an extremely strict experimental context. At the underlying text generation base (Base LLM) level, all three uniformly adopt DeepSeek-V3 to ensure that performance differences in the dynamic evaluation are strictly limited to the system control architecture (state machine and gating mechanism vs. pure prompts) itself, excluding variable disturbances caused by the inherent capabilities of the base model.

Meanwhile, to avoid the self-preference bias that might arise from interactions between homologous models, the State-Blind Seeker Simulator in this study is driven by another independent base model, Qwen-Plus. All systems face this identical heterogeneous simulated seeker and read the same case narrative anchors (such as initial emotion, dilemma type) and maximum interaction turn constraints. This heterogeneous model design effectively circumvents the self-preference bias that may arise from homologous model interactions, further guaranteeing the objectivity of the dynamic evaluation.

\subsection{Process-Oriented Metrics}
\label{sec:metrics}
Unlike traditional evaluations that rely on text similarity, this paper designs a set of Process-oriented Metrics that quantify interaction timing and closed-loop completion rates, using LLM-as-Judge to conduct stage-based scoring for each dynamic trajectory \cite{zheng2023judging}. To ensure objectivity, these metrics are strictly defined as observable "Event Judgments," rather than subjective evaluations of language style or empathy quality: (1) \textbf{Assessment (Asse)} evaluates whether the system establishes a comprehensive understanding of the user's presenting situation and core conflict before transitioning; (2) \textbf{Education (Edu)} examines whether it delivers psychoeducation by surfacing cognitive patterns and offering actionable guidance; (3) \textbf{Intervention (Int)} requires both confirming a practice goal and completing a behavioral simulation with reflection for a strict closed-loop; and (4) \textbf{Homework (Hw)} measures the establishment of a concrete self-directed continuation plan. Furthermore, we construct a Resistance Hard-Negative Stress Test of 50 extreme high-resistance cases to evaluate boundary discipline under adversarial conditions, detailed in Section~\ref{sec:stress}.

\subsection{Main Dynamic Evaluation Results}

\begin{table}[t]
\caption{Main Dynamic Evaluation Results (300 Samples)}
\label{tab:main-dynamic-results}
\centering
\small
\setlength{\tabcolsep}{4pt} 
\begin{tabular}{lcccc}
\toprule
Architecture & Asse & Edu & Int & Hw \\
\midrule
Baseline (Prompt-only) & 1.00 & 0.44 & 0.71 & 0.21 \\
MiddleBaseline (Stage-prompt) & 1.00 & 0.70 & 0.51 & 0.36 \\
LEKIA (Situated Architecture) & 1.00 & 0.92 & 0.81 & 0.85 \\
\bottomrule
\end{tabular}
\end{table}

As shown in Table~\ref{tab:main-dynamic-results}, all systems saturate the shallow Assessment stage, indicating that LLMs can reliably manage basic listening and empathy relying on context understanding. However, in deeper intervention stages requiring strict process control, LEKIA significantly outperforms the baselines. The gap is most pronounced in the strict closed-loop metrics: LEKIA achieves 0.81 in \textit{int}, outperforming both Baseline (0.71) and MiddleBaseline (0.51). Similarly, in \textit{hw}, LEKIA scores 0.85, significantly higher than the baselines (0.21 and 0.36, respectively). Averaged across all four stages, LEKIA achieves a mean completion rate of 0.895, representing an absolute improvement of approximately 31\% over the prompt-only Baseline and 25\% over MiddleBaseline.

MiddleBaseline's performance provides a critical empirical contrast, with its int score (0.51) notably falling behind the pure Baseline (0.71). This counter-intuitive result reveals the structural risk of partial stage awareness without completion grounding. Stage prompts give the model a local sense of its current stage, but without an external completion check, the model implicitly treats its own intervention-style output—framing exercises, inviting reflection—as evidence of progress, causing the stage to stall in substantively incomplete loops.
Compounding this, MiddleBaseline's prompt-based cooldown introduces a contradictory constraint: the model is simultaneously instructed to "conduct exercises" and "suppress advancement." The pure Baseline, unconstrained by explicit stage logic, follows conversational momentum—a cruder strategy, but one that avoids the self-canceling signals introduced by partial control. This confirms that partial control is worse than no control: it suppresses the LLM's natural progression instincts without providing reliable process guarantees. LEKIA avoids this failure mode precisely because stage advancement depends entirely on externally maintained completion flags rather than the model's own self-assessed output—an advantage rooted not in finer prompt engineering, but in externalizing process control into a continuously updatable state structure that provides hard guarantees for stage completion beyond text generation.

\begin{table}[t]
\caption{Inter-Rater Reliability: LLM vs. Human Annotators}
\label{tab:judge-reliability}
\centering
\small
\begin{tabular}{lc}
\toprule
Rater Pair & Cohen's $\kappa$ \\
\midrule
Annotator A  vs.\ Judge      & 0.8081 \\
Annotator B  vs.\ Judge & 0.7218 \\
Annotator A vs.\ Annotator B  & 0.7271 \\
\bottomrule
\end{tabular}
\end{table}

\subsection{Judge Reliability Audit}
As show in table~\ref{tab:judge-reliability} The process-oriented metrics in this paper are observable event judgments, rather than subjective language style or empathy quality scorings. Nevertheless, to systematically verify the reliability of LLM-as-Judge on such structured event judgments, we blindly sampled 50 evaluation anchors from the main experiment logs. A computer science expert (Annotator A) and a domain expert with a background in psychology and neuroscience (Annotator B) independently scored the same samples and compared results with the Judge's automatic output.
The results showed that the Cohen's Kappa~\cite{cohen1960kappa}coefficient between Annotator A and the Judge was 0.8081, between Annotator B and the Judge was 0.7218, and the Kappa between the two human annotators was 0.7271. These results confirm that the LLM-as-Judge reliably tracks stage boundaries and closed-loop completion as structured event judgments, achieving agreement levels consistent with human expert consensus.

\subsection{Robustness Analysis across Different Distress Types}
To test the robustness of the main experiment's conclusions across different scenarios, we conducted a stratified analysis of stage metrics by distress type based on the existing 300 dynamic samples. Seven categories with sufficient sample sizes (n$\geq$20) are included in the main text for comparison, while the remaining categories are listed in the appendix. During the assessment stage, all systems were close to saturation across all main types, so it is not the main focus of expansion here.

\begin{table}[t]
\caption{Cross-Type Robustness (Baseline / MiddleBaseline / LEKIA)}
\label{tab:cross-type-snapshots}
\centering
\small
\setlength{\tabcolsep}{3pt} 
\begin{tabular}{lccc} %
\toprule
Distress Type & edu & int & hw \\
\midrule
School bullying   & .56/.74/.91 & .65/.41/.91 & .51/.40/.76 \\
Job crisis        & .42/.68/.94 & .78/.56/.83 & .44/.40/.88 \\
Breakup           & .33/.77/.92 & .72/.57/.70 & .61/.38/.90 \\
Friend conflict   & .34/.64/.89 & .67/.39/.78 & .58/.25/.94 \\
Academic pressure & .39/.69/.89 & .67/.65/.75 & .32/.33/.87 \\
Depression        & .45/.55/.90 & .70/.30/.77 & .45/.50/.83 \\
Sleep problems    & .48/.82/.95 & .80/.56/.89 & .46/.60/1.0 \\
\bottomrule
\end{tabular}
\end{table}

As shown in Table~\ref{tab:cross-type-snapshots}, the cross-type advantage of edu is the most stable, with LEKIA leading the two baseline groups in all 7 types of distress, indicating that the cognitive layer's support for the psychoeducation closed-loop has consistent cross-scenario robustness.

int presents some heterogeneity. LEKIA maintains a stable advantage over MiddleBaseline, but the gap with Baseline is limited in job crisis (0.83 vs 0.78) and breakup with partner (0.70 vs 0.72)---the narrative structures of these two distress types are relatively linear, and Baseline's aggressive progression style makes it easier to achieve a superficial intervention closed-loop locally, but this local score does not extend to subsequent stages.

The distribution of hw best reflects the architectural differences: LEKIA reaches above 0.76 in all 7 categories, while MiddleBaseline is only between 0.25 and 0.60 in most types, and even lower than Baseline in problems with friends (0.25) and breakup with partner (0.38). This consistent pattern confirms that LEKIA's advantage is architectural rather than scenario-specific.

\subsection{Resistance Hard-Negative Stress Test}
\label{sec:stress}
In addition to the main experiment, a stress test utilizing 50 selected extreme high-resistance cases further revealed the systems' baseline capabilities in preventing false progression and maintaining intervention resilience. Handling resistance involves a complete timeline containing immediate concession, cooldown waiting, and opportunistic retries. To this end, we jointly evaluated the Immediate Adherence Rate, Cooldown Violation Rate, and Eventual Re-offer Rate. In the turn where the seeker explicitly refuses or hesitates (Turn-0), LEKIA achieved an Immediate Adherence Rate of up to 0.84. In contrast, Baseline was only 0.3, making it highly prone to disregarding willingness and forcefully preaching. MiddleBaseline's Immediate Adherence Rate (0.42) was only slightly higher than Baseline's (0.3), showing limited improvement. More alarmingly, its cooldown violation rate (0.48) not only failed to improve, but was significantly higher than Baseline's (0.28)---this indicates that the continuous pressure from stage prompts somewhat exacerbated the model's impulsivity during the cooldown period, making it more inclined to immediately initiate repeated offers.

\begin{table}[t]
\caption{Resistance Hard-Negative Stress Test (50 Cases)}
\label{tab:stress-test}
\centering
\begin{tabular}{lccc}
\toprule
Metric & Baseline & MiddleBaseline & LEKIA \\
\midrule
Immediate Adherence Rate & 0.30 & 0.42 & 0.84 \\
Cooldown Violation Rate & 0.28 & 0.48 & 0.00 \\
Eventual Re-offer Rate & 0.30 & 0.80 & 0.66 \\
\bottomrule
\end{tabular}
\end{table}

Furthermore, preventing false progression does not mean the system should fall into "passive empathic stagnation"; an effective process-oriented support system should retain bounded proactivity. In this long-term contest, Baseline's eventual re-offer rate was only 0.30, indicating limited re-engagement after encountering resistance. Although MiddleBaseline's eventual re-offer rate reached 0.80, this occurred alongside its high cooldown violation rate of (\textasciitilde{}0.48). In contrast, LEKIA, while producing no cooldown violations in this stress-test setting, achieved an eventual re-offer rate of up to 0.66. This indicates that, under controlled simulation, the system can resume the intervention trajectory while respecting cooldown constraints and updated signals of user readiness.

Together, these three evaluations form a converging chain of evidence: process completion under normal conditions, boundary discipline under resistance, and cross-scenario consistency. For process-oriented emotional support, external state and 
dual-gating are not enhancements to generation quality, but prerequisites for maintaining intervention safety.

\section{Discussion and Limitations}
\subsection{The Value of External State for Process Control}
The results demonstrate that explicit external state provides an effective mechanism for process control in the evaluated setting: without it, even architecturally aware systems like 
MiddleBaseline suffer from stage drift and boundary violations that 
prompt engineering alone cannot prevent.

Traditional affective-computing evaluation has focused heavily on 
utterance quality, which can obscure process-level failures in 
long-horizon interaction. The Static-to-Dynamic protocol addresses 
this gap by shifting emphasis from local linguistic quality to 
global multi-turn process control, offering a reusable path to more standardized safety evaluation of LLM-based emotional support 
systems. Separating process control from language generation also 
reveals that backbone model value is not determined solely by 
reasoning strength, but by its balance among naturalness, 
controllability, and process discipline.

The MiddleBaseline results expose a subtler risk: linguistic fluency 
can actively mask process failure. A system may produce responses 
that sound empathetic while silently drifting from its intervention 
trajectory---a failure mode invisible to both users and standard 
metrics. This ``fluency illusion'' is particularly dangerous in 
high-stakes support scenarios, and points to the need for 
process-oriented safety metrics as a distinct evaluation dimension.

A deeper implication concerns the status of user consent in dialogue 
system design. Prevailing approaches treat consent as a soft 
alignment target---an emergent property of careful training rather 
than a structural guarantee. LEKIA's dual-gating mechanism 
operationalizes a different principle: that a person's expressed 
hesitation is not an obstacle to be managed, but a boundary to be 
unconditionally respected. By encoding the User Gate as a hard 
precondition and cooldown as an enforced state, the architecture 
embeds human dignity into its control logic---the system cannot 
advance against a person's will not because it has been trained to 
hesitate, but because its structure makes doing so impossible. This 
shift from consent as alignment objective to consent as architectural 
invariant points toward a design philosophy in which user autonomy 
is not approximated but guaranteed.

\subsection{Limitations and Future Work}
This study has several clear limitations, which also point to important directions for future work.

First, the reliability of the dynamic evaluation remains bounded. The main experiment relies on LLM-as-judge for process-level scoring. Although expert review of 50 samples showed relatively high agreement between automatic scores and human judgment, this should not be taken to imply the absence of systematic bias in automatic evaluation. Future evaluation in real-world support settings should therefore include larger-scale double-blind human review as well as direct assessment in genuine interactive environments.

Second, the cross-lingual robustness of the architecture has been validated only in English and Chinese deployments. While the language-agnostic stage progress flags provide a structural foundation for broader multilingual extension, whether the same architectural invariance holds under typologically distinct languages—such as those with fundamentally different pragmatic norms or affective expression conventions—remains an open empirical question for future work.

Third, although this work includes a front-end prototype, it has not yet been validated through real-user studies. The current findings are derived primarily from dynamic simulation-based evaluation and prototype-level deployment, and therefore cannot fully establish the system's acceptability, naturalness, or long-term usability in real interaction settings. Future work should include systematic user studies with real participants and front-end environments to evaluate practical effectiveness.

\section{Conclusion}
This paper presents LEKIA 2.0, a situated LLM architecture with an explicit separation between the Cognitive Layer and the Executive Layer, designed to address stateless drift in process-oriented emotional support. By combining external state modeling with a strictly gated progression mechanism, the system successfully maintains stable situational awareness and user consent boundaries throughout continuous interaction. Furthermore, the proposed Static-to-Dynamic evaluation protocol provides a reusable framework to systematically quantify multi-turn process-control capabilities that are difficult to evaluate through static text alone.

\section*{Author Contributions}
\textbf{Boning Zhao}: Conceptualization, Methodology, Software (Core Architecture \& Backend System), Data Curation, Investigation, Project Administration, and Writing -- Original Draft. \\
\textbf{Yutong Hu}: Conceptualization and Methodology (psychology-informed design of the architecture and evaluation protocol), Data Curation (case construction and annotation), and Writing -- Review \& Editing.  \\
\textbf{Xinnuo Li}: Software (Frontend Live2D \& Audio Interface), Visualization, and Validation.

\section*{Ethical Considerations and Data Availability}
This work is motivated by a commitment to responsible development 
of AI-assisted mental health tools. All evaluation data are 
synthetically constructed via domain-expert-guided augmentation 
of existing public corpora; no real user data, clinical records, 
or personally identifiable information were collected or used. We are mindful that emotional support systems operate in contexts 
of heightened vulnerability. LEKIA's architectural choices---explicit 
consent boundaries, cooldown enforcement, and stage-constrained 
progression---reflect an ethical stance: people seeking support 
deserve systems whose protective behaviors are structurally 
guaranteed rather than statistically likely. Our goal is not to automate therapy, but to extend evidence-based supportive practices to those who may otherwise lack access to consistent and boundary-respecting care.

All datasets, model code, and front-end prototype will be made 
publicly available upon acceptance of the final version.

\bibliographystyle{splncs04}
\bibliography{refs}

\end{document}